\documentclass[runningheads]{llncs}

\usepackage{eccv}

\usepackage{eccvabbrv}

\usepackage{graphicx}
\usepackage{booktabs}

\usepackage[accsupp]{axessibility}  

\usepackage{array}
\usepackage{multirow}    
\usepackage[table]{xcolor}
\usepackage[most]{tcolorbox}
\usepackage{marvosym}
\usepackage{textcomp}
\usepackage[normalem]{ulem}
\usepackage{caption}

\definecolor{myLightGreen}{HTML}{E6F4EA}  
\definecolor{myLightRed}{HTML}{FDECEC}   
\definecolor{myLightOrange}{HTML}{FFF3E0}  
\definecolor{navyblue}{RGB}{0,0,128}
\newcommand{\equalcont}{\textsuperscript{*}}

\usepackage{hyperref}
\usepackage{wrapfig}
\usepackage{orcidlink}

\begin{document}

\title{MonoSR: Open-Vocabulary Spatial Reasoning on Monocular Images}

\titlerunning{MonoSR}

\newcommand{\corrauthor}{\textsuperscript{\Letter}}

\author{
Qirui Wang\equalcont\inst{1} \and
Jingyi He\equalcont\inst{1} \and
Yining Pan\inst{2,3} \and
Si Yong Yeo\inst{4} \and
Xulei Yang\corrauthor\inst{2} \and
Shijie Li\corrauthor\inst{2}
}
\authorrunning{Q. Wang et al.}

\institute{
Technical University of Munich, Germany
\and
A*STAR Institute of Advanced Intelligence and Computing, Singapore
\and
Singapore University of Technology and Design, Singapore
\and
Nanyang Technological University, Singapore
\\[0.4em]
\email{\{li\_shijie, yang\_xulei\}@a-star.edu.sg}
}
\maketitle

\begingroup
\renewcommand{\thefootnote}{}
\footnotetext{
* Equal contribution. \qquad
\Letter\ Corresponding author.
}
\endgroup

\begin{figure*}[t]
  \centering
  \includegraphics[width=\textwidth]{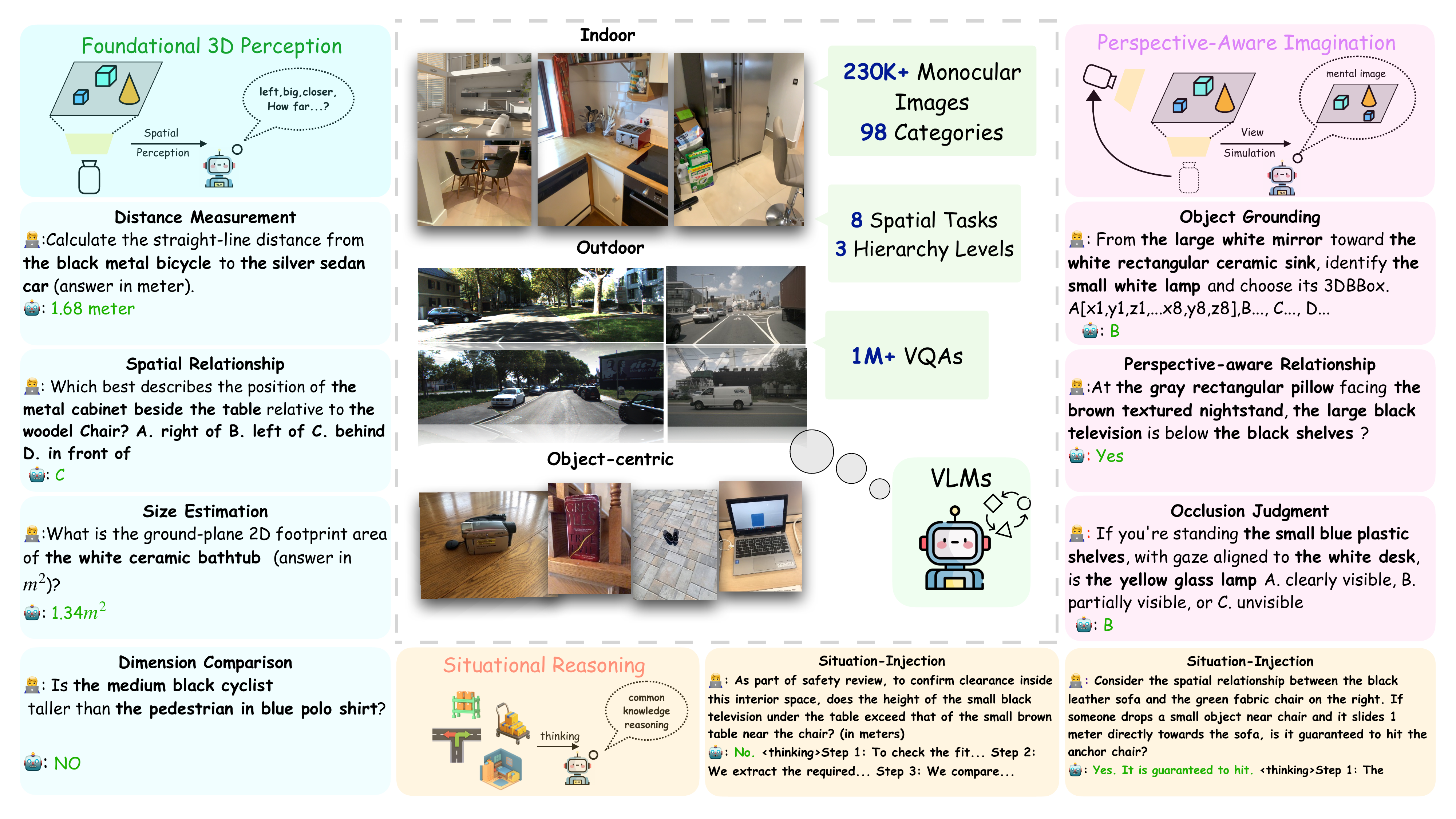}
  \caption{
        Overview of the proposed MonoSR dataset, which spans three levels of spatial reasoning—Foundational Perception, Perspective-Aware Imagination, and Situational Reasoning—across diverse indoor, outdoor, and object-centric scenes. The dataset contains over 1 million spatial VQA samples, providing comprehensive coverage for open-world monocular spatial reasoning. 
        }
  \label{fig:overview}
\end{figure*}

\begin{abstract}
Spatial reasoning from a single RGB image, including the inference of distances, sizes, and inter-object relationships without depth sensors or multiple views, remains a critical blind spot for current Vision Language Models (VLMs). Existing benchmarks either rely on multi-view video sequences that expose explicit geometric cues, 
or are confined to indoor environments too small for model training. 
To close this gap, we introduce MonoSR, a large-scale dataset for 
open-world monocular spatial reasoning comprising over 1M QA pairs 
from 230K images spanning indoor, outdoor, and object-centric 
domains across 98 semantic categories. Every QA pair is guaranteed 
answerable from a single RGB image via a four-stage observability 
filter validated by human audit. 
Tasks are organized into three cognitive levels: foundational 
perception, perspective-aware imagination, and situational 
reasoning, supporting both large-scale training and fine-grained 
evaluation. Comprehensive benchmarking of state-of-the-art open- 
and closed-source VLMs reveals consistent limitations across all 
three domains, with object-centric metric estimation emerging as 
the most challenging frontier. We further conduct a systematic 
auxiliary information study by injecting scene context, 2D visual 
prompts, and 3D bounding boxes, allowing us to quantify the geometric gap that 
future monocular perception modules must close and provide 
actionable design guidelines. \textbf{Our project page is available at this\href{https://7rwang.github.io/MonoSR/}{ link}.} 
\end{abstract}

\section{Introduction}
\label{sec:intro}

Vision-Language Models (VLMs)\cite{Bai2023QwenVL,GeminiTeam2023Gemini, Touvron2023LLaMA,Achiam2023GPT4,Liu2023VisualInstructionTuning} have recently achieved significant progress in general visual understanding and cross-modal reasoning\cite{Radford2021CLIP,Li2023BLIP2,ren2024grounded, liu2024grounding,wu2026lavit,he2026surgonair}. These models align visual information into the semantic space of a pretrained LLM, excelling at high-level recognition and language-grounded tasks. Yet a single photograph is enough for a human to judge whether a chair fits through a doorway, estimate how far a car is from a pedestrian, or reason about what lies just outside the frame, but this effortless monocular spatial reasoning remains a critical blind spot for current VLMs\cite{Li2025ViewSpatialBench}. This ability, referred to as Spatial Reasoning~(SR), is essential for physical-world applications such as Embodied AI.

While a variety of datasets have been proposed to advance SR research\cite{Wald2020Learning3DSceneGraphs,GuoSURDS,Tian2025NuScenesSpatialQA, Achlioptas2020ReferIt3D,Ma2023SQA3D,Marino2019OKVQA}, these efforts predominantly rely on multi-view images, video sequences, or point clouds, where explicit geometric cues are readily available. As a result, the monocular setting, where such cues are absent, remains largely underexplored. Furthermore, existing datasets are mostly confined to indoor environments and focus on low-level perception, leaving higher-order spatial imagination and open-world generalization as open challenges. Consequently, monocular spatial reasoning in open-world scenarios remains fundamentally underserved as both a training resource and an evaluation framework.

To address these limitations, we present MonoSR, a large-scale dataset 
for open-world monocular spatial reasoning with guaranteed single-view 
answerability. An overview of MonoSR is shown in Fig.~\ref{fig:overview}. 
MonoSR takes a single RGB image as input, without relying on the explicit geometric cues present in 
multi-view or video-based settings. Beyond input modality, MonoSR spans 
indoor, outdoor, and object-centric domains across 98 semantic categories, 
supporting both large-scale training and open-world evaluation. Tasks are 
organized into three hierarchical cognitive levels: foundational perception, 
perspective-aware imagination, and situational reasoning~\cite{zhang2025flatland,
Lee2025PerspectiveAware,Collins2024BuildingMachines}.

We further investigate the role of auxiliary information in monocular 
spatial reasoning. Unlike multi-view methods that extract spatial cues 
from multiple observations, monocular reasoning must operate without 
explicit geometric recovery. To provide design guidelines for future 
monocular perception modules, we fine-tune Qwen-2.5VL-3B~\cite{Bai2023QwenVL} 
under varied input configurations—scene information, 2D visual prompts, 
3D bounding boxes, and their combinations—and analyze how each signal 
contributes to spatial reasoning performance.

Our main contributions are as follows:
\begin{itemize}
\item We introduce MonoSR, a large-scale dataset for open-world 
monocular spatial reasoning featuring over 1M QA pairs from 230K 
images spanning indoor, outdoor, and object-centric domains across 
98 semantic categories, supporting both training and open-world evaluation.

\item We propose a Single-View Observability Guarantee mechanism 
that systematically ensures every QA pair is answerable from a single 
RGB image alone, via a four-stage filtering pipeline validated by 
human audit.

\item We conduct a systematic investigation into auxiliary information 
for monocular spatial reasoning, offering practical design guidelines 
for future monocular perception modules such as metric depth estimation 
and monocular 3D detection. Comprehensive evaluation of state-of-the-art 
VLMs further reveals consistent limitations in this challenging setting.
\end{itemize}

\section{Related Work}

\begin{figure*}[!t]
  \centering
  \includegraphics[width=\textwidth,trim=220 240 220 250,clip]{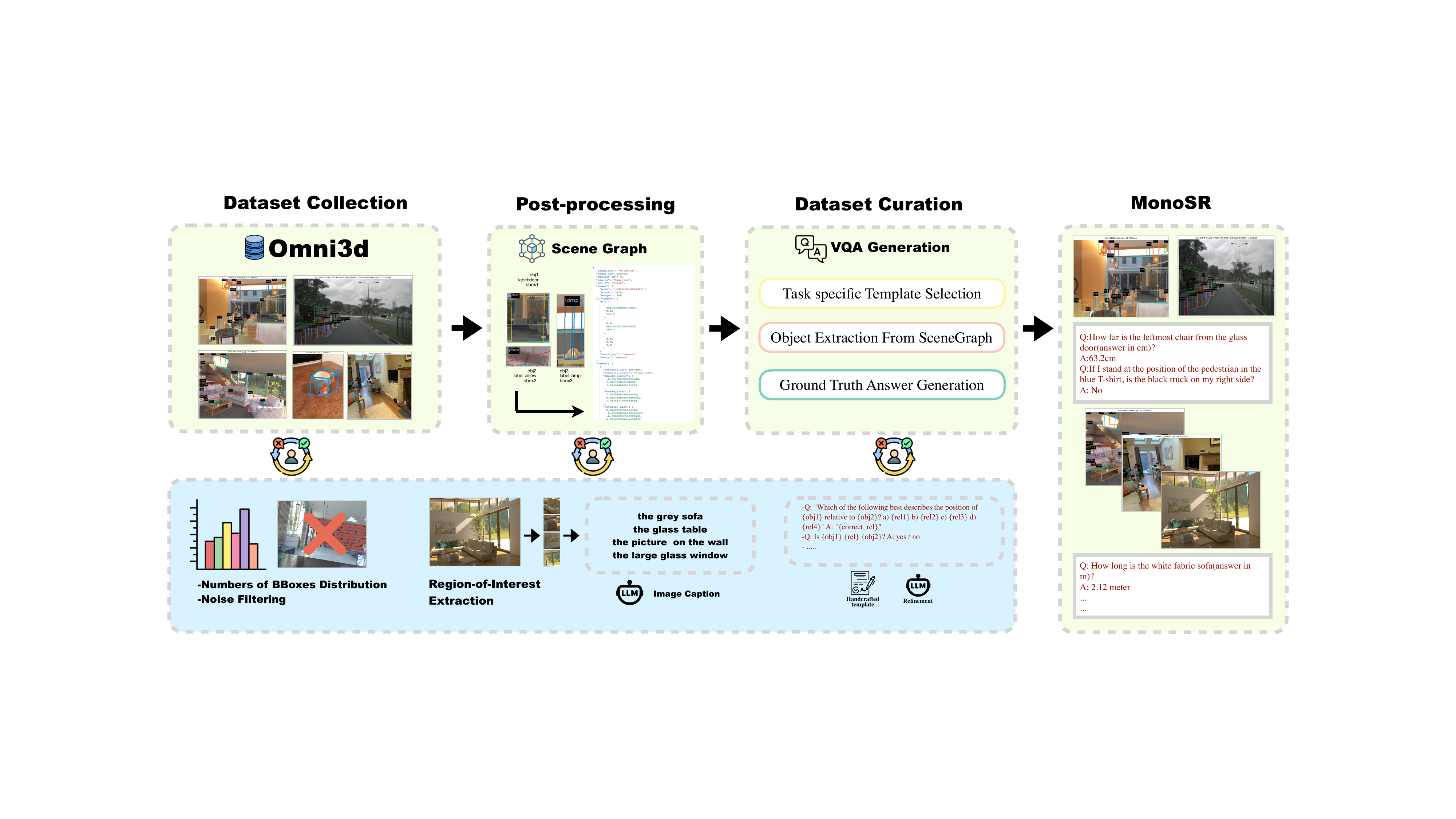}
  \caption{Demonstration of MonoSR curation pipeline}
  \label{fig:pipeline}
\end{figure*}

\subsection{3D Spatial Reasoning Datasets}
The emergence of new datasets has significantly advanced 3D spatial reasoning, enabling models to better interpret spatial relationships in 3D environments\cite{Chen2024SpatialVLM,Cheng2024SpatialRGPT,szymanska2024space3dbench,Zhu_2024_ScanReason,jiang2025vknowu}. More recent efforts focus on evaluating 3D reasoning capabilities directly within VLM architectures, with representative diagnostic benchmarks such as VSR\cite{Liu2023VisualSpatialReasoning}, 3DSRBench\cite{Ma2025ThreeDSRBench}, SEED-Bench\cite{Li2024SeedBench} SpatialEval\cite{Wang2024PictureWorthThousandWords}, and VSI-Bench\cite{Yang2025ThinkingInSpace}. These benchmarks are typically built upon existing 3D reconstruction datasets that primarily cover indoor environments and provide video sequences for each scene, resulting in multi-view images as the default input setting.
Similarly, several datasets have been constructed to endow models with 3D spatial reasoning ability, for example, ScanQA\cite{Azuma2022ScanQA}, SQA3D\cite{Ma2023SQA3D}, and ReferIt3D\cite{Achlioptas2020ReferIt3D}, which incorporate depth and geometry to support question answering within reconstructed indoor scenes. More recent datasets\cite{Fan2025VLM3R,wu2025spatial,cai2025spatialbot} expand the number of 3D scenes and task types; however, they remain limited to indoor domains and are not optimized for monocular inputs.
Unlike these benchmarks, MonoSR is designed with monocular-first constraints; details are provided in Sec.~\ref{sec:monosr}.

\subsection{Spatial Reasoning Methods}
Current Vision–Language Models (VLMs) can be broadly categorized into general-purpose and specialized models. General-purpose VLMs\cite{Bai2023QwenVL,Liu2023VisualInstructionTuning,Achiam2023GPT4,Touvron2023LLaMA} demonstrate strong capability in 2D understanding but struggle with 3D spatial tasks. This limitation arises from their pre-training on predominantly 2D image–text corpora, which lack the geometric grounding required for accurate monocular spatial inference.
To overcome this limitation, specialized 3D-aware approaches have been developed. Many of these methods project explicit 3D features, such as point-cloud representations, into the language model\cite{Hong2023ThreeDLLM,Xu2024PointLLM,Qi2024ShapeLLM}. More recently, perceiving and reasoning about 3D environments using only 2D observations has gained increasing attention.
SeeGround~\cite{li2025seeground} bridges 3D visual grounding and 2D VLMs by rendering query-aligned views and enriching them with spatial descriptions, while perception-aware monocular reasoning methods~\cite{cheng2026perception} introduce explicit object-centric visual grounding for spatial reasoning from a single image.
Other recent systems, such as SpatialReasoner~\cite{zheng2025spatialreasoner}, further explore active perception and tool-augmented reasoning for large-scale 3D scene understanding.
In parallel, several 3D-aware VLM methods~\cite{Fan2025VLM3R,wu2025spatial,meng2025phymagic} feed multi-view images into the model to obtain spatial predictions.
However, their effectiveness is restricted by existing datasets, which predominantly contain indoor scenes and are not optimized for monocular inputs, a far more common scenario in real-world applications.
Notably, specialized spatial reasoning methods such as SpatialVLM~\cite{Chen2024SpatialVLM} and LLMI3D~\cite{yang2026llmi3d} are originally designed for multi-view or depth-augmented inputs.
Beyond metric geometry, prior work has also explored functional affordance grounding and memory-based reference resolution~\cite{wang2026grounding}, highlighting the broader role of spatial grounding in embodied contexts.
This motivates our situational reasoning tasks, which connect spatial scene structure with realistic operational contexts.

\section{MonoSR}
\label{sec:monosr}


\subsection{Dataset Curation \& Annotations }\label{Dataset Curation & Annotations}

MonoSR is built upon Omni3D \cite{brazil2023omni3d}, a large-scale dataset designed for open-vocabulary monocular 3D object detection across diverse scenarios. In MonoSR, the input images are sourced directly from Omni3D, while the answers are derived from the corresponding ground-truth 3D bounding boxes. The overall dataset curation and annotation pipeline is illustrated in Fig. \ref{fig:pipeline}.

\noindent\textbf{Design Principles.}
Realizing these goals introduces three concrete engineering challenges. First, spatial quantities must be defined relative to the camera coordinate frame rather than an absolute world frame, ensuring that answers remain well-defined under the single-view constraint. Second, unanswerable questions arising from occlusion, truncation, or annotation noise must be systematically identified and excluded via a multi-stage observability filter. Third, generating ground-truth answers for perspective-aware imagination tasks requires rigid geometric transformation rather than direct image inspection. The following subsections describe the mechanisms we design to address each challenge.

\noindent\textbf{Cognition-Inspired Task Design}
We design eight tasks across three hierarchical levels, mirroring how humans perceive and reason about the 3D world~\cite{Li2025SpatialLadder,Chen2025PerceptionBeforeReasoning}.
(1) \textit{Foundational 3D Perception} evaluates fundamental geometric and spatial properties, including spatial relationships~(SR), size estimation~(Size), distance measurement~(Dist), and dimension comparison~(Dim).
(2) \textit{Perspective-Aware Imagination} assesses the model's ability to answer questions from unobserved viewpoints, requiring implicit perspective imagination and spatial consistency~\cite{Zhang2025SpinBenchWorkshop}. This is related to novel-view understanding and synthesis~\cite{li2025valid}, but is evaluated through language-based spatial reasoning rather than image generation. This level comprises occlusion judgment~(OJ), perspective-aware relationship~(PR), and object grounding~(OG).
(3) \textit{Situational Reasoning} examines complex inference capabilities by simulating real-world scenarios that require integrating visual scene information with common knowledge. To produce natural and coherent questions, our QA generation prompt assigns the LLM an expert persona (\eg, a safety inspector or a roboticist) and instructs it to embed a \textit{motivation clause} grounding the question in a realistic operational context. An example is shown below:

\begin{tcolorbox}[
    title={Prompt},
    colback=black!5,      
    colframe=black!90,  
    coltitle=white,      
    fonttitle=\bfseries, 
    arc=2mm,              
    boxrule=0.6pt,  
      left=1mm,      
    right=1mm, 
      before skip=2.5pt,  
    after skip=2.5pt,
    top=0.5mm,
    bottom=0.5mm,
]
\small 
\texttt{"}You are an expert annotator for 3D scene understanding and spatial reasoning. Your job is to rewrite the original question into a realistic professional scenario, including a brief motivation and the required metric about the target object ...\texttt{"}
\end{tcolorbox}

Each task supports Yes/No, Multiple-Choice, and Numerical question formats to comprehensively assess spatial understanding at different response granularities.

\noindent\textbf{Scene Pre-processing and Representation.} 
To generate semantically rich and geometrically accurate QA pairs from raw Omni3D scenes, we design a three-stage preprocessing pipeline. First, we apply a rigorous filtering procedure to select high-quality monocular images that satisfy our criteria for scene complexity and visual clarity. Second, to minimize referential ambiguity, we generate fine-grained descriptive captions for all salient objects within each selected scene. Finally, we parse the captioned objects and their 3D relationships to construct a comprehensive scene graph, which serves as the structured ground-truth foundation for subsequent QA generation.

\textbf{QA Data Generation.} Before QA generation, MonoSR performs quality control on the 3D annotations by filtering unreliable object boxes, including those with implausible dimensions (e.g., larger than 20,m), near-zero volume, or substantial discrepancies between their 2D projections and 3D footprints (greater than 30\%). Using the resulting high-quality scene graphs, we adopt a two-stage hybrid generation strategy. In the first stage, all answers are deterministically derived from 3D ground-truth annotations: answers for \textit{Foundational 3D Perception} tasks are computed directly from object coordinates, while those for \textit{Perspective-Aware Imagination} are obtained through rigid transformations and geometric consistency checks. Questions are paired with these verified answers via handcrafted templates, ensuring systematic coverage across all task categories. In the second stage, to mitigate the rigidity and template bias inherent in the initial set~\cite{Cheng2024SpatialRGPT}, an LLM paraphrases and diversifies the question text while preserving the original ground-truth answer, yielding more natural linguistic styles without compromising geometric fidelity.

\begin{wrapfigure}[18]{r}{0.5\columnwidth} 
  \centering
  \includegraphics[
    width=\linewidth,
    height=0.23\textheight,
    keepaspectratio,
    trim=400 1 350 1,
    clip
]{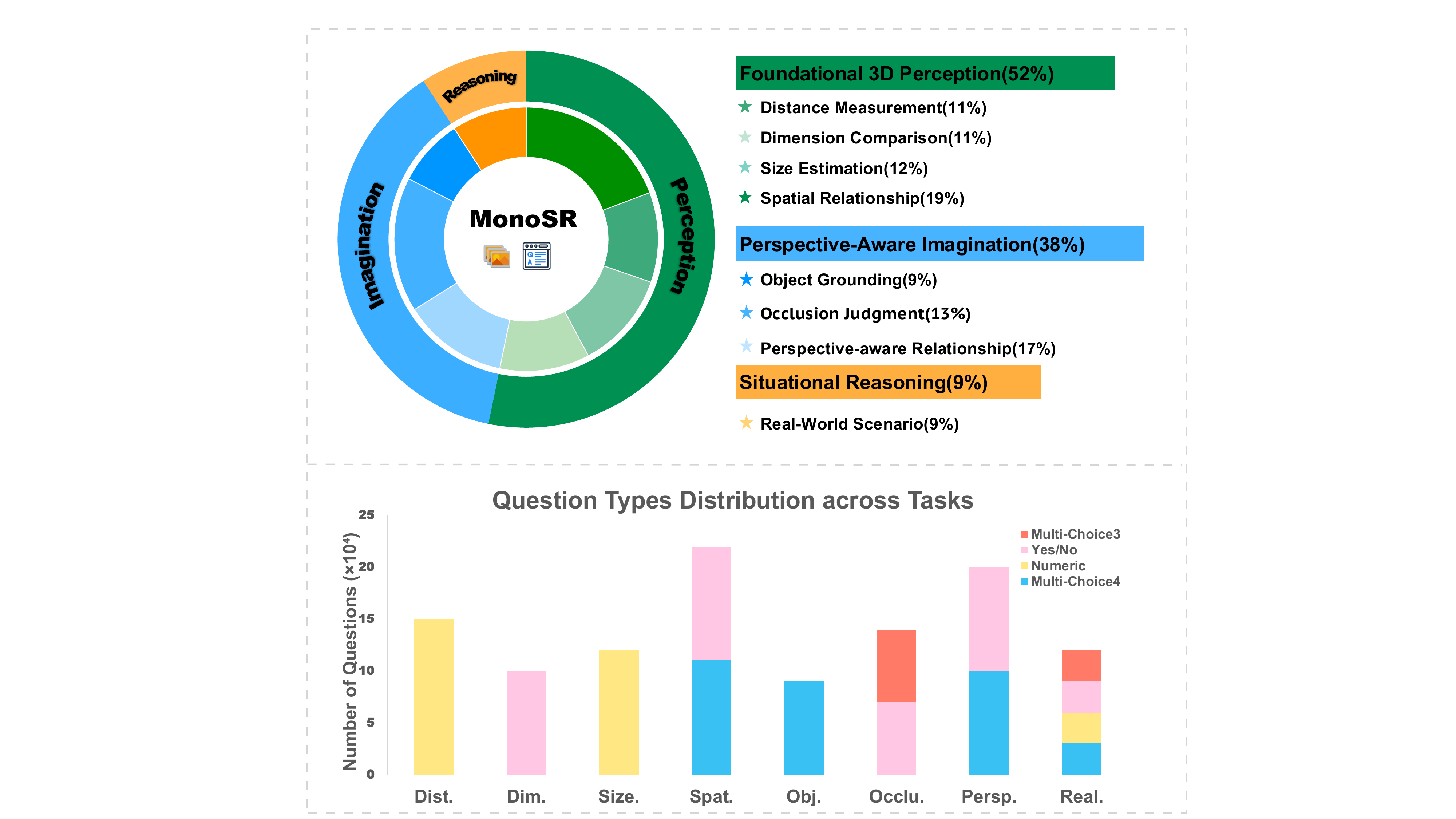}
  \caption{Dataset composition of MonoSR across hierarchical reasoning levels and domains. \textbf{Top:} Distribution of tasks across three reasoning levels. \textbf{Bottom:} Breakdown of question formats.}
  \label{fig:statistics}
\end{wrapfigure}

\noindent\textbf{Single-View Observability Guarantee}
A core requirement of MonoSR is that every QA pair must be answerable from the single RGB image alone. We provide this guarantee through a four-stage pipeline spanning both scene preprocessing (stages i–iii) and post-generation QA validation (stage iv):
(i)~\textit{instance visibility}: truncation $<$ 0.4, occlusion $<$ 0.6, and projected area $>$ 400\,px$^2$;
(ii)~\textit{3D annotation quality}: degenerate or geometrically inconsistent bounding boxes are discarded;
(iii)~\textit{scene complexity}: each scene must contain 3--40 valid instances, balancing richness against label noise;
(iv)~\textit{QA ambiguity}: questions with visually indistinguishable answers (\eg, depth-ambiguous pairs with $<$20\,px separation) are removed.
 Human audit confirms that residual unanswerable questions account for $<$1.4\% of the retained set.

\subsection{Statistics \& Analysis}
\label{Statistics & Analysis}

\textbf{Dataset Statistics.} An overview of MonoSR statistics is shown in Fig.~\ref{fig:statistics}.
MonoSR comprises over 1.02 million QA pairs constructed from 
more than 230K high-resolution monocular images, covering indoor, outdoor, 
and object-centric domains derived from Omni3D. Each image contains on 
average 20 annotated objects across 98 semantic categories, which is 
substantially larger than those in previous indoor-based spatial reasoning 
datasets~\cite{Fan2025VLM3R,wu2025spatial}.

As illustrated in Fig.~\ref{fig:statistics}, the dataset follows a 
balanced hierarchical composition: 52\% of QA pairs belong to 
\textit{Foundational 3D Perception}, 38\% to \textit{Perspective-Aware 
Imagination}, and 10\% to \textit{Situational Reasoning}. To further 
prevent models from exploiting dataset priors, we enforce fine-grained 
balance across multiple dimensions: Yes/No polarity is set to 50/50, 
MC option positions are randomized, spatial relation categories are 
constrained to within 10\% relative deviation, and answer types are 
equalized across tasks. More details are provided in the 
\textit{Supplemental Materials}.

\noindent\textbf{Quality Validation.}
To ensure reliability, we conducted both automated and human validation. 
Geometric consistency checks against 3D annotations achieve exceptionally 
high correctness for perception-level tasks. 
Furthermore, 8 independent annotators manually audited 10\% of the validation 
set QA pairs and reported an inter-annotator agreement of $\mathcal{ACC}=98.6\%$. 
Error analysis reveals three primary failure modes: visual ambiguity 
(\eg, heavily occluded or extremely small objects), boundary conditions 
(\eg, objects near the image border), and rare annotation noise induced 
by LLM paraphrasing. The overall error rate is low, and the strict 
single-view filtering strategy effectively suppresses the majority of 
such cases prior to inclusion.

\noindent\textbf{Comparison with Existing Benchmarks.}
Tab.~\ref{tab:dataset_comparison} situates MonoSR among representative 
spatial reasoning benchmarks along five axes: scene coverage, input 
modality, dataset scale, and support for model training.

Three advantages of MonoSR are evident. 
\textit{First}, MonoSR is the only dataset to cover all three scene domains 
simultaneously—indoor, outdoor, and object-centric—whereas all prior 
benchmarks are confined to indoor reconstructions, limiting their 
applicability to real-world deployment scenarios such as autonomous driving 
and robotic manipulation.
\textit{Second}, MonoSR enforces \textit{explicit single-view answerability}: 
every QA pair is generated and filtered to be answerable from a single RGB 
image via viewpoint-aware grounding, with visually ambiguous or 
non-observable cases systematically removed. 
Existing benchmarks~\cite{Ma2023SQA3D,Azuma2022ScanQA,Yang2025ThinkingInSpace} 
do not impose this constraint and instead rely on multi-view video sequences 
or depth inputs that expose explicit geometric cues unavailable under 
monocular conditions.
\textit{Third}, at over 1M QA pairs from 230K images, MonoSR is the 
\textit{only monocular spatial reasoning dataset that simultaneously supports 
large-scale training and open-world evaluation}. 
Existing monocular-compatible benchmarks such as SPHERE ($\sim$6K) and 
OmniSpatial (8.4K) are purely diagnostic and too small for model training, 
preventing the community from fairly developing and evaluating monocular 
spatial reasoning models at scale.

\begin{table}[t]
\centering
\caption{
    Comparison of MonoSR with representative spatial reasoning benchmarks.
    \textbf{Mono-only}: explicitly designed for single monocular RGB input,
    without relying on multi-view, video, or depth cues.
    \textbf{Train}: large enough to support model fine-tuning in addition
    to evaluation. \textbf{Obj.}: object-centric scene coverage.
}
\label{tab:dataset_comparison}
\resizebox{\columnwidth}{!}{
\setlength{\tabcolsep}{5pt}
\renewcommand{\arraystretch}{1.2}
\begin{tabular}{l | c c c | r r | c c}
\toprule
\textbf{Dataset}
    & \textbf{Indoor}
    & \textbf{Outdoor}
    & \textbf{Obj.}
    & \textbf{\#Images}
    & \textbf{\#QA Pairs}
    & \textbf{Mono-only}
    & \textbf{Train} \\
\midrule
ScanQA~\cite{Azuma2022ScanQA}
    & \checkmark & $\times$ & $\times$
    & 800        & 41K
    & $\times$   & \checkmark \\

SQA3D~\cite{Ma2023SQA3D}
    & \checkmark & $\times$ & $\times$
    & 650        & 33K
    & $\times$   & \checkmark \\

EmbSpatial-Bench~\cite{du2024embspatial}
    & \checkmark & $\times$ & $\times$
    & $\sim$2K   & 3.6K
    & $\times$   & $\times$ \\

SPHERE~\cite{zhang2024sphere}
    & \checkmark & $\times$ & $\times$
    & $\sim$3K   & $\sim$6K
    & \checkmark & $\times$ \\

OmniSpatial~\cite{jia2025omnispatial}
    & \checkmark & $\times$ & $\times$
    & 6.5K       & 8.4K
    & \checkmark & $\times$ \\

VSI-Bench~\cite{Yang2025ThinkingInSpace}
    & \checkmark & $\times$ & $\times$
    & 288        & 5K
    & $\times$   & $\times$ \\

RoboSpatial\cite{song2025robospatial}
    & \checkmark & $\times$ & $\times$
    & 1M         & 3M
    & $\times$   & \checkmark \\

SpatialBot\cite{cai2025spatialbot}
    & \checkmark & \checkmark & $\times$
    & 743K       & 743K
    & $\times$   & \checkmark \\

InternSpatial\cite{deng2025internspatial}
    & \checkmark & \checkmark & \checkmark
    & --         & 12M
    & $\times$   & \checkmark \\

Ego3D-Bench\cite{gholami2025spatial}
    & $\times$   & \checkmark & $\times$
    & --         & $\sim$9K
    & $\times$   & $\times$ \\
\midrule
\rowcolor{myLightGreen}
\textbf{MonoSR (Ours)}
    & \checkmark & \checkmark & \checkmark
    & \textbf{230K}
    & \textbf{1M+}
    & \checkmark & \checkmark \\
\bottomrule
\end{tabular}
}
\end{table}

\section{Auxiliary Information Evaluation}
\label{sec:aux}

Most recent spatial reasoning methods are designed for multi-view inputs, 
which implicitly provide 3D spatial cues via plug-in extraction 
modules~\cite{Fan2025VLM3R,wu2025spatial}. Such strategies cannot be 
directly applied to the monocular setting, where only a single image is 
available and explicit 3D structure cannot be reliably recovered. Aided 
by MonoSR, we conduct a comprehensive investigation into the impact of 
auxiliary information on monocular spatial reasoning, aiming to provide 
guidance for future method design. Importantly, we treat 3D bounding 
boxes as an \textit{upper-bound oracle}: not a practical monocular 
solution, but a controlled probe to quantify the geometric information 
gap that future monocular perception modules (\eg, metric depth estimation,
monocular 3D detection, and monocular scene completion~\cite{li2025global}) must close.

\begin{wrapfigure}{r}{0.5\columnwidth}
  \centering
  \includegraphics[width=\linewidth]{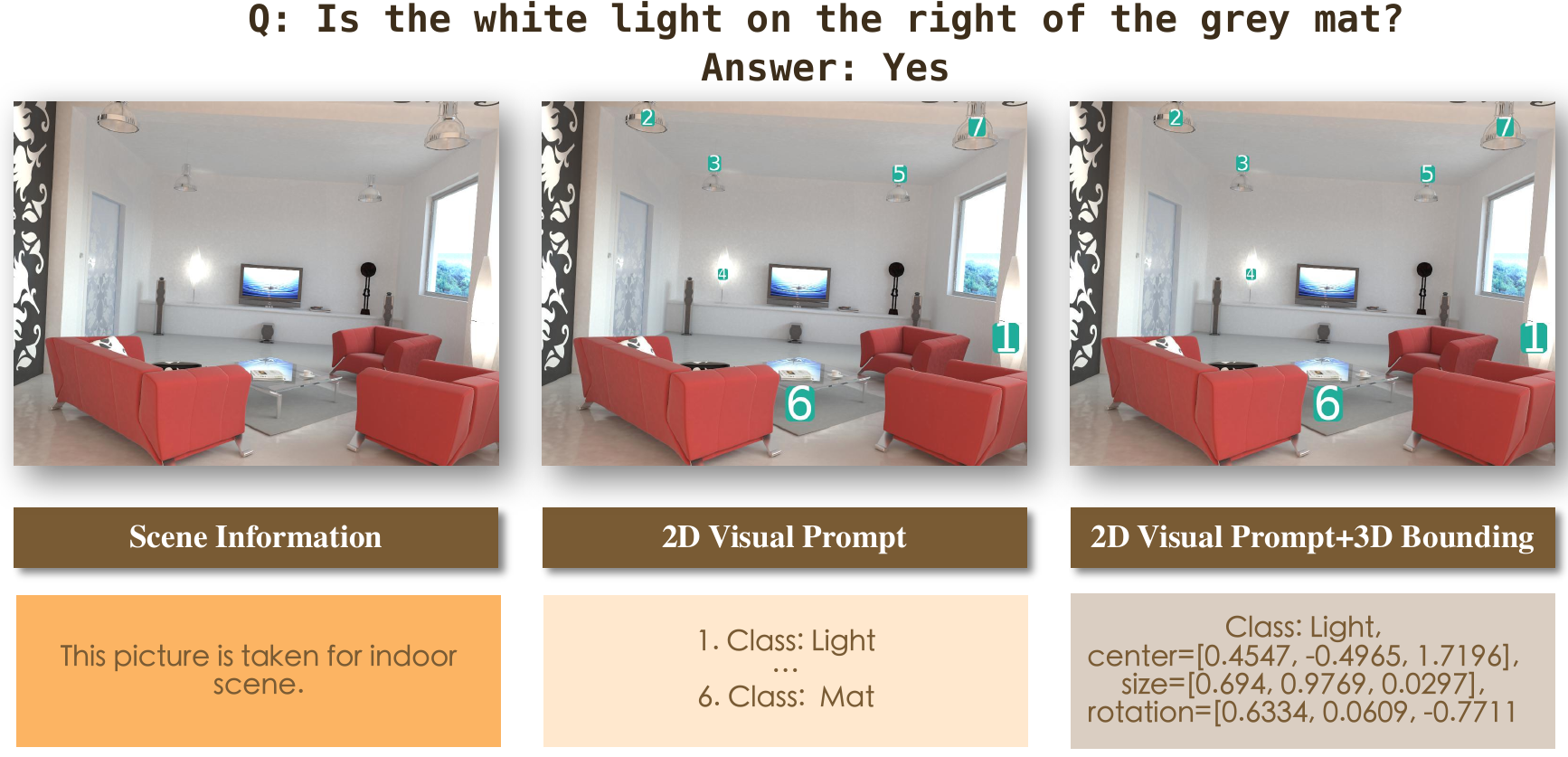}
  \caption{Illustration of the three auxiliary information types investigated in this work: Scene Information (SI), 2D Visual Prompts (2D VP), and 3D Bounding Boxes (3D Bbox).}
  \label{fig:Prompt}
  \vspace{-4mm}
\end{wrapfigure}

Specifically, we explicitly inject the following auxiliary signals into 
the input, either individually or in combination, and evaluate how each 
contributes to the final performance. As illustrated in 
Fig.~\ref{fig:Prompt}, the information investigated includes:
\begin{itemize}
    \item Scene Information (SI): Specifies whether input image 
    is captured in an indoor, outdoor, or object-centric setting, 
    and is provided on textual side.
    \item 2D Visual Prompt (2D VP): Explicitly marks detected objects 
    on the input image to highlight their 2D coordinates, and is 
    provided through the visual (image) input.
    \item 3D Bounding Box (3D Bbox): Represents object using 
     3D center coordinates, spatial dimensions, and 
    orientation, and is provided on  textual side.
\end{itemize}
Visualizations of three auxiliary information types are provided 
in Fig.~\ref{fig:Prompt}. Additional details regarding experimental 
setup and results are presented in Sec.~\ref{sec:exp}.

\section{Experiments}
\label{sec:exp}

\begin{table}[!ht]
    \captionsetup{type=table}
    \caption{Benchmarking open \& closed-source methods. \uline{Some unreasonable question types are filtered out depending on the scenario.} Dark blue and orange indicate the best result among all models, while light colours indicate the second best result.}
    \centering
    \resizebox{\textwidth}{!}{
    \large 
    \begin{tabular}{r|c|ccccc|ccccc|cccccccccc}
    & &
    \rotatebox{75}{SR(MC4)} &
    \rotatebox{75}{SR(Y/N)} &
    \rotatebox{75}{Dist(Num.)} &
    \rotatebox{75}{Dim(Y/N)} &
    \rotatebox{75}{Size (Num.)} &
    \rotatebox{75}{OJ(MC3)} &
    \rotatebox{75}{OJ(Y/N)} &
    \rotatebox{75}{PR(MC4)} &
    \rotatebox{75}{PR(Y/N)} &
    \rotatebox{75}{OG(MC4)} &
    \rotatebox{75}{High SR(MC4)} &
    \rotatebox{75}{High SR(Y/N)} &
    \rotatebox{75}{High Dist(Num.)} &
    \rotatebox{75}{High Dim(Y/N)} &
    \rotatebox{75}{High Size (Num.)} &
    \rotatebox{75}{High OJ(MC3)} &
    \rotatebox{75}{High OJ(Y/N)} &
    \rotatebox{75}{High PR(MC4)} &
    \rotatebox{75}{High PR(Y/N)} &
    \rotatebox{75}{High OG(MC4)}
    \\
    Methods & Overall &
    \multicolumn{5}{c}{\cellcolor{red!10}Low-level} &
    \multicolumn{5}{c}{\cellcolor{orange!10}Mid-level} &
    \multicolumn{10}{c}{\cellcolor{yellow!10}High-level} \\
    \hline

    \rowcolor{navyblue!10}
    \multicolumn{22}{c}{\textbf{\textit{Indoor Scene}}} \\
    
\rowcolor{navyblue!5}
\multicolumn{1}{l|}{\textcolor{black}{\textit{Baseline}}} & & & & & & & & & & & & & & & & & & & & &\\
Qwen-2.5-VL-3B 
     & 0.353 
    & 0.312 & 0.495 & 0.063 & 0.440 & 0.074  
    & 0.286 & 0.164 & 0.247 & 0.491 & 0.241 
    & 0.330 & 0.330 & 0.025 & 0.423 & 0.041 &
      0.269 & 0.520 & 0.256 & 0.441 & 0.207 \\
\hline

\rowcolor{navyblue!5}
\multicolumn{1}{l|}{\textit{Open-source Models}} & & & & & & & & & & & & & & & & & & & & &\\
LLaVA-OneVision-72B        
     & 0.334 
    & 0.252 & 0.510 & 0.011 & 0.619 & 0.018 
    & 0.271 & 0.528 & 0.254 & 0.491 & 0.270
    & 0.340 & 0.490 & 0.000  & 0.470 & 0.064 &
    0.290 & \cellcolor{navyblue!50}0.660 & 0.240 & 0.490 & 0.240 \\

Qwen-2.5-VL-72B     
     & 0.357 
    & 0.361 & 0.493 & 0.068 & \cellcolor{navyblue!20}0.651 & 0.123 
    & 0.085 & 0.562 & 0.317 & 0.496 & 0.279
    & 0.427 & \cellcolor{navyblue!20}0.520 & 0.021 & \cellcolor{navyblue!20}0.571 & 0.137 &
      0.220 & 0.620 & \cellcolor{navyblue!20}0.285 & 0.412 & 0.280 \\

Intern-VL-3-78B     
     & 0.363 
    & 0.336 & \cellcolor{navyblue!20}0.521 & 0.063 & 0.585 & 0.091 
    & 0.273 & \cellcolor{navyblue!20}0.639 & 0.310 & 0.482 & 0.280
    & 0.430 & 0.482 & 0.041 & 0.520 & 0.114 &
      0.290 & 0.630 & 0.260 & 0.423 & 0.340 \\

SpatialVLM    
     & \cellcolor{orange!50}0.511 
    & \cellcolor{navyblue!50}0.529 & 0.476 & \cellcolor{navyblue!50}0.505 & 0.589 & \cellcolor{navyblue!50}0.421 
    & \cellcolor{navyblue!50}0.444 & 0.554 & \cellcolor{navyblue!50}0.585 & 0.447 & \cellcolor{navyblue!20}0.535
    & \cellcolor{navyblue!50}0.598 & 0.462 & 0.457 & 0.442 & 0.464 &
      \cellcolor{navyblue!50}0.616 & 0.582 & \cellcolor{navyblue!50}0.454 & \cellcolor{navyblue!50}0.583 & \cellcolor{navyblue!20}0.475 \\
\hline

\rowcolor{navyblue!5}
\multicolumn{1}{l|}{\textit{Closed-source Models}} & & & & & & & & & & & & & & & & & & & & &\\
ChatGPT-4 
    & 0.301 
    & 0.335 & 0.487 & 0.008 & 0.603 & 0.035 
    & 0.096 & 0.111 & 0.267 & \cellcolor{navyblue!20}0.499 & 0.279
    & 0.353 & 0.512 & 0.013 & \cellcolor{navyblue!50}0.615 & 0.013 &
      0.240 & \cellcolor{navyblue!20}0.647 & 0.133 & 0.464 & 0.188 \\

Gemini-2.5-Pro 
    & 0.379 
    & 0.385 & 0.493 & \cellcolor{navyblue!20}0.089 & \cellcolor{navyblue!50}0.677 & 0.136 
    & 0.276 & 0.161 & 0.348 & 0.484 & 0.443
    & 0.409 & 0.515 & 0.002 & 0.500 & 0.100 &
      0.228 & 0.557 & 0.212 & 0.412 & 0.464 \\

Gemini-2.5-Flash 
    & 0.380 
    & 0.339 & 0.499 & 0.081 & 0.576 & \cellcolor{navyblue!20}0.140 
    & \cellcolor{navyblue!20}0.316 & 0.205 & \cellcolor{navyblue!20}0.398 & \cellcolor{navyblue!50}0.515 & 0.259
    & 0.410 & 0.490 & 0.031 & 0.540 & 0.121 &
      \cellcolor{navyblue!20}0.320 & 0.600 & 0.231 & 0.470 & 0.310 \\

Gemini-3-Flash-Preview
    & 0.373
    & 0.397 & \cellcolor{navyblue!50}0.523 & 0.000 & 0.606 & 0.000
    & 0.303 & 0.616 & 0.339 & 0.489 & 0.310
    & 0.280 & 0.500 & \cellcolor{navyblue!20}0.505 & 0.310 & \cellcolor{navyblue!50}0.670
    & 0.250 & 0.480 & 0.280 & \cellcolor{navyblue!20}0.520 & 0.360 \\

GPT-5.2
    & \cellcolor{orange!15}0.399
    & \cellcolor{navyblue!20}0.479 & 0.514 & 0.000 & 0.642 & 0.000
    & 0.255 & \cellcolor{navyblue!50}0.685 & 0.280 & 0.482 & \cellcolor{navyblue!50}0.594
    & \cellcolor{navyblue!20}0.462 & \cellcolor{navyblue!50}0.547 & \cellcolor{navyblue!50}0.581 & 0.400 & \cellcolor{navyblue!20}0.539
    & 0.260 & 0.520 & 0.168 & 0.436 & \cellcolor{navyblue!50}0.557 \\
  
    \hline\hline

    
\rowcolor{navyblue!10}
\multicolumn{22}{c}{\textbf{\textit{Outdoor Scene}}} \\

\rowcolor{navyblue!5}
\multicolumn{1}{l|}{\textit{Baseline}} & & & & & & & & & & & & & & & & & & & & &\\
Qwen-2.5-VL-3B 
     & 0.301 
    & 0.385 & 0.474 & 0.039 & $\times$ & $\times$ 
    & $\times$ & $\times$ & 0.241 & 0.504 & 0.250 
    & 0.304 & 0.430 & 0.023 & $\times$ & $\times$ &
      $\times$ & $\times$ & 0.280 & 0.430 & 0.261 \\
\hline

\rowcolor{navyblue!5}
\multicolumn{1}{l|}{\textit{Open-source Models}} & & & & & & & & & & & & & & & & & & & & &\\

LLaVA-OneVision-72B & 0.328
    & 0.262 & 0.510 & 0.000 
    & $\times$ & $\times$
    & $\times$ & $\times$
    & 0.254 & \cellcolor{navyblue!20}0.522 & 0.272
    & 0.360 & \cellcolor{navyblue!50}0.600
    & 0.000 & $\times$ & $\times$
    & $\times$ & $\times$
    & 0.330 & \cellcolor{navyblue!50}0.530 & 0.300 \\

Qwen-2.5-VL-72B & 0.390
    & 0.439 & 0.538 & 0.068 
    & $\times$ & $\times$
    & $\times$ & $\times$
    & 0.368 & 0.464 & 0.328
    & \cellcolor{navyblue!20}0.460 & 0.550
    & \cellcolor{navyblue!20}0.089 & $\times$ & $\times$
    & $\times$ & $\times$
    & 0.380 & 0.450 & 0.354 \\

Intern-VL-3-78B & 0.345
    & 0.414 & 0.521 & 0.019 
    & $\times$ & $\times$
    & $\times$ & $\times$
    & 0.396 & \cellcolor{navyblue!50}0.527 & 0.253
    & 0.350 & 0.560 
    & 0.023 & $\times$ & $\times$
    & $\times$ & $\times$
    & 0.260 & 0.420 & 0.310 \\

SpatialVLM & 0.383
    & \cellcolor{navyblue!50}0.505 & \cellcolor{navyblue!50}0.607 & \cellcolor{navyblue!50}0.582 
    & $\times$ & $\times$
    & $\times$ & $\times$
    & \cellcolor{navyblue!50}0.486 & 0.454 & \cellcolor{navyblue!20}0.493
    & 0.420 & 0.469 
    & \cellcolor{navyblue!50}0.578 & $\times$ & $\times$
    & $\times$ & $\times$
    & 0.422 & 0.539 & \cellcolor{navyblue!50}0.540 \\
\hline

\rowcolor{navyblue!5}
\multicolumn{1}{l|}{\textit{Closed-source Models}} & & & & & & & & & & & & & & & & & & & & &\\

ChatGPT-4 & 0.337
    & 0.426 & 0.523 & 0.012 
    & $\times$ & $\times$
    & $\times$ & $\times$
    & 0.426 & 0.488 & 0.276
    & 0.346 & 0.520
    & 0.000 & $\times$ & $\times$
    & $\times$ & $\times$
    & 0.293 & 0.432 & 0.286 \\

Gemini-2.5-Pro & \cellcolor{orange!20}0.394
    & \cellcolor{navyblue!20}0.448 & 0.519 & 0.082 
    & $\times$ & $\times$
    & $\times$ & $\times$
    & 0.384 & 0.485 & 0.380
    & 0.435 & 0.470
    & 0.014 & $\times$ & $\times$
    & $\times$ & $\times$
    & 0.355 & 0.469 & 0.459 \\

Gemini-2.5-Flash & 0.378
    & 0.393 & 0.530 & 0.055 
    & $\times$ & $\times$
    & $\times$ & $\times$
    & \cellcolor{navyblue!20}0.446 & 0.482 & 0.361
    & 0.374 & \cellcolor{navyblue!20}0.581
    & 0.014 & $\times$ & $\times$
    & $\times$ & $\times$
    & 0.347 & 0.440 & 0.343 \\

Gemini-3-Flash-Preview
    & \cellcolor{orange!15}0.392
    & 0.423 & \cellcolor{navyblue!20}0.545 & 0.000
    & $\times$ & $\times$
    & $\times$ & $\times$
    & 0.386 & 0.465 & 0.481
    & \cellcolor{navyblue!20}0.460 & 0.550
    & 0.084 & $\times$ & $\times$
    & $\times$ & $\times$
    & 0.410 & 0.500 & 0.410 \\

GPT-5.2
    & \cellcolor{orange!50}0.415
    & 0.422 & 0.516 & 0.000
    & $\times$ & $\times$
    & $\times$ & $\times$
    & 0.353 & 0.491 & \cellcolor{navyblue!50}0.588
    & \cellcolor{navyblue!50}0.568 & 0.475
    & 0.012 & $\times$ & $\times$
    & $\times$ & $\times$
    & \cellcolor{navyblue!50}0.470 & \cellcolor{navyblue!20}0.520 & \cellcolor{navyblue!20}0.536 \\

    \hline\hline

\rowcolor{navyblue!10}
\multicolumn{22}{c}{\textbf{\textit{Object-Centric Scene}}} \\

\rowcolor{navyblue!5}
\multicolumn{1}{l|}{\textit{Baseline}} & & & & & & & & & & & & & & & & & & & & &\\
Qwen-2.5-VL-3B 
    & 0.345
    & $\times$ & $\times$ & 0.064 & $\times$ & 0.057 
    & $\times$ & $\times$
    & $\times$ & $\times$ & $\times$
    & $\times$ & $\times$ & 0.043 & $\times$ & 0.045 
    & $\times$ & $\times$ & $\times$ & $\times$ & $\times$ \\
\hline

\rowcolor{navyblue!5}
\multicolumn{1}{l|}{\textit{Open-source Models}} & & & & & & & & & & & & & & & & & & & & &\\

LLaVA-OneVision-72B 
    & 0.087
    & $\times$ & $\times$ & 0.104 & $\times$ & 0.014 
    & $\times$ & $\times$
    & $\times$ & $\times$ & $\times$
    & $\times$ & $\times$ & 0.012 & $\times$ & 0.020 
    & $\times$ & $\times$ & $\times$ & $\times$ & $\times$ \\

Qwen-2.5-VL-72B 
    & 0.395
    & $\times$ & $\times$ & 0.194 & $\times$ & 0.148
    & $\times$ & $\times$
    & $\times$ & $\times$ & $\times$
    & $\times$ & $\times$ & \cellcolor{navyblue!20}0.122 & $\times$ & \cellcolor{navyblue!20}0.157 
    & $\times$ & $\times$ & $\times$ & $\times$ & $\times$ \\

Intern-VL-3-78B 
    & 0.275
    & $\times$ & $\times$ & 0.148 & $\times$ & 0.101 
    & $\times$ & $\times$
    & $\times$ & $\times$ & $\times$
    & $\times$ & $\times$ & 0.024 & $\times$ & 0.141 
    & $\times$ & $\times$ & $\times$ & $\times$ & $\times$ \\

SpatialVLM 
    & \cellcolor{orange!50}0.618
    & $\times$ & $\times$ & \cellcolor{navyblue!50}0.622 & $\times$ & \cellcolor{navyblue!50}0.569 
    & $\times$ & $\times$
    & $\times$ & $\times$ & $\times$
    & $\times$ & $\times$ & \cellcolor{navyblue!50}0.598 & $\times$ & \cellcolor{navyblue!50}0.682 
    & $\times$ & $\times$ & $\times$ & $\times$ & $\times$ \\
\hline

\rowcolor{navyblue!5}
\multicolumn{1}{l|}{\textit{Closed-source Models}} & & & & & & & & & & & & & & & & & & & & &\\

ChatGPT-4 
    & 0.069
    & $\times$ & $\times$ & 0.021 & $\times$ & 0.061 
    & $\times$ & $\times$
    & $\times$ & $\times$ & $\times$
    & $\times$ & $\times$ & 0.000 & $\times$ & 0.021 
    & $\times$ & $\times$ & $\times$ & $\times$ & $\times$ \\

Gemini-2.5-Pro 
    & \cellcolor{orange!20}0.440
    & $\times$ & $\times$ & \cellcolor{navyblue!20}0.305 & $\times$ & 0.168 
    & $\times$ & $\times$
    & $\times$ & $\times$ & $\times$
    & $\times$ & $\times$ & 0.081 & $\times$ & 0.062 
    & $\times$ & $\times$ & $\times$ & $\times$ & $\times$ \\

Gemini-2.5-Flash 
    & 0.433
    & $\times$ & $\times$ & 0.160 & $\times$ & \cellcolor{navyblue!20}0.235 
    & $\times$ & $\times$
    & $\times$ & $\times$ & $\times$
    & $\times$ & $\times$ & 0.073 & $\times$ & 0.151 
    & $\times$ & $\times$ & $\times$ & $\times$ & $\times$ \\

Gemini-3-Flash-Preview
    & 0.130
    & $\times$ & $\times$ & 0.145 & $\times$ & 0.118
    & $\times$ & $\times$
    & $\times$ & $\times$ & $\times$
    & $\times$ & $\times$ & 0.122 & $\times$ & 0.135
    & $\times$ & $\times$ & $\times$ & $\times$ & $\times$ \\

GPT-5.2
    & 0.087
    & $\times$ & $\times$ & 0.094 & $\times$ & 0.081
    & $\times$ & $\times$
    & $\times$ & $\times$ & $\times$
    & $\times$ & $\times$ & 0.076 & $\times$ & 0.097
    & $\times$ & $\times$ & $\times$ & $\times$ & $\times$ \\

    \end{tabular}
 }
      
    \label{table:benchmarking}
\end{table}

\subsection{Evaluation Setup}

\textbf{Benchmark Models.} We evaluate diverse representative vision-language models (VLMs) that cover both open-source and closed-source paradigms. The open-source group includes \textit{LLaVA-OneVision-72B}\cite{Liu2023VisualInstructionTuning}, \textit{Qwen-2.5-VL-72B-Instruct}\cite{Bai2023QwenVL},  \textit{Intern-VL-3-78B}\cite{zhu2025internvl3}, and \textit{SpatialVLM}\cite{Chen2024SpatialVLM}. The closed-source group consists of \textit{ChatGPT-4}\cite{Achiam2023GPT4}, \textit{Gemini-2.5-Pro}, and \textit{Gemini-2.5-Flash} \cite{GeminiTeam2023Gemini}. Additionally, we include \textit{Qwen-2.5-VL-3B}\cite{Bai2023QwenVL} as a lightweight baseline to serve as the performance anchor across all spatial domains. 

\textbf{Evaluation Domains.} MonoSR includes three distinct spatial domains: Indoor, Outdoor, and Object-Centric, and along a hierarchical reasoning ladder consisting of low-level, mid-level, and high-level dimensions. This setup enables a fine-grained analysis of a model’s ability to generalize monocular spatial understanding across complexity levels and visual domains.  One thing to note is that we filter out certain questions under specific scenarios. For example, spatial-relationship questions are not meaningful for object-centric images, as in most cases only a single object is visible.

\textbf{Metric Design.} For each question type, we design an appropriate evaluation protocol. Specifically, we use accuracy to evaluate Yes/No and Multiple-Choice questions. 
For Numerical questions, we account for potential scale drift by introducing an adaptive threshold. If the discrepancy between the prediction and the ground truth falls within this threshold, the answer is considered correct. Specifically, the threshold is set to 10\% of the ground-truth value, and a prediction is deemed correct only if it satisfies $|\frac{\hat{d} - d}{d}| < 0.1 $ where $d$ is ground truth value and $\hat{d}$ is model prediction.

\noindent\textbf{Metric Sensitivity Analysis.}
To validate the robustness of the 10\% threshold, we additionally 
evaluate all models under a stricter 5\% relative-error criterion. 
As shown in Tab.~\ref{tab:threshold}, model rankings remain 
consistent across both thresholds, confirming that the chosen 
tolerance does not inflate results or alter comparative conclusions. 
Under the 5\% criterion, performance drops substantially across 
all models, indicating that the task remains far from saturation 
and that the 10\% threshold reflects a perceptually reasonable 
accuracy level rather than an artificially lenient standard.

\begin{table}[t]
\centering
\caption{
    Sensitivity analysis of the numerical evaluation threshold. 
    Model rankings remain consistent between 10\% and 5\% 
    relative-error criteria, confirming robustness of the metric.
}
\label{tab:threshold}
\resizebox{\columnwidth}{!}{
\setlength{\tabcolsep}{5pt}
\renewcommand{\arraystretch}{1.2}
\begin{tabular}{l | cc | cc | cc}
\toprule
& \multicolumn{2}{c|}{\textbf{Indoor}} 
& \multicolumn{2}{c|}{\textbf{Outdoor}} 
& \multicolumn{2}{c}{\textbf{Object-Centric}} \\
\textbf{Model} 
& \textbf{10\%} & \textbf{5\%} 
& \textbf{10\%} & \textbf{5\%} 
& \textbf{10\%} & \textbf{5\%} \\
\midrule
Qwen-2.5-VL-3B       & 0.063 & 0.031 & 0.039 & 0.018 & 0.064 & 0.029 \\
LLaVA-OV-72B      & 0.011 & 0.004 & 0.000 & 0.000 & 0.104 & 0.047 \\
Qwen-2.5-VL-72B   & 0.068 & 0.034 & 0.068 & 0.033 & 0.194 & 0.091 \\
InternVL-3-78B    & 0.063 & 0.029 & 0.019 & 0.008 & 0.148 & 0.065 \\
ChatGPT-4         & 0.008 & 0.003 & 0.012 & 0.005 & 0.021 & 0.009 \\
Gemini-2.5-Pro    & 0.089 & 0.041 & 0.082 & 0.039 & 0.305 & 0.143 \\
Gemini-2.5-Flash  & 0.081 & 0.038 & 0.055 & 0.025 & 0.160 & 0.073 \\
\bottomrule
\end{tabular}
}
\end{table}

\subsection{Benchmark Open- and Closed-Source Methods}

Tab.~\ref{table:benchmarking} presents a detailed breakdown of model performance under indoor, outdoor, and object-centric scenarios, revealing several consistent patterns across low-, mid-, and high-level spatial reasoning tasks.

\textbf{Indoor Scene.}
\textit{SpatialVLM} achieves the strongest overall open-source 
performance (0.511), excelling on low-level geometric tasks 
(SR(MC4) = 0.529, Dist = 0.505) and mid-level relation reasoning 
(PR(MC4) = 0.585). General-purpose models such as \textit{InternVL-3-78B} 
and \textit{Qwen-2.5-VL-72B} remain competitive on qualitative 
relation tasks (Mid-OJ(Y/N)) but struggle with numerical estimation. 
Among closed-source models, \textit{Gemini-2.5-Pro} and 
\textit{Gemini-2.5-Flash} outperform ChatGPT-4 across most categories, 
particularly on high-level compositional tasks such as High-OG(MC4).

\begin{figure*}[t]
  \centering
  \includegraphics[width=\textwidth,
                   trim=1 230 1 280,clip]{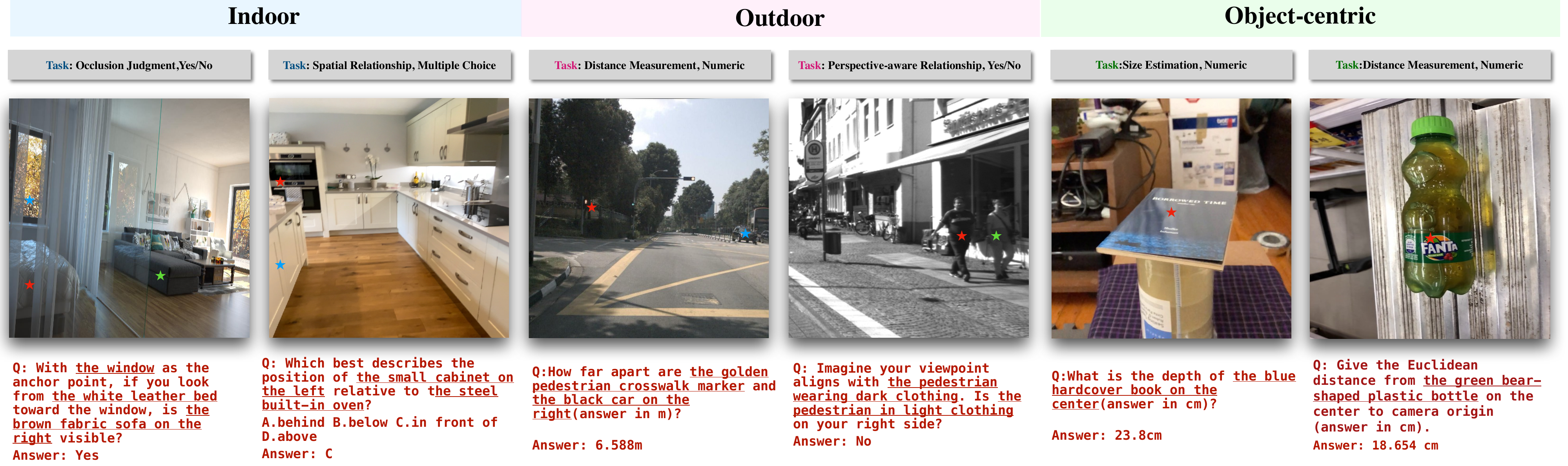}
  \caption{\textbf{Dataset visualization.} Examples from indoor, outdoor, and object-centric scenarios covering multiple spatial reasoning tasks with paired questions and ground-truth answers.}
  \label{fig:visualization}
\end{figure*}

\textbf{Outdoor Scene.}
Performance patterns broadly mirror the indoor setting. \textit{SpatialVLM} 
again leads on low-level metric estimation (Dist = 0.582, SR(MC4) = 0.505) 
and mid-level relation reasoning (PR(MC4) = 0.486), benefiting from its 
geometry-aware training. Among open-source models, the strongest mid-level interaction result (PR(Y/N) = 0.527) 
is achieved by \textit{InternVL-3-78B}.
Closed-source models \textit{Gemini-2.5-Pro} and \textit{Gemini-2.5-Flash} 
retain their advantage on high-level reasoning, particularly High-OG(MC4).

\textbf{Object-Centric Scene.}
This setting is the most challenging for most models, as it demands 
fine-grained metric reasoning (size and distance) from a single object 
view. \textit{SpatialVLM} achieves a clear advantage (Overall = 0.618, 
Dist = 0.622, Size = 0.569), while general-purpose open-source models 
collapse significantly—\textit{LLaVA-OneVision-72B} drops to 0.087. 
Among closed-source models, \textit{Gemini-2.5-Pro} (0.440) and 
\textit{Gemini-2.5-Flash} (0.433) show moderate capability but remain 
well below SpatialVLM on quantitative tasks.

\begin{wraptable}[24]{r}{0.6\textwidth} 
\centering
\caption{Performance on low-frequency categories across three levels: high-level, mid-level, and low-level.}
\small
\resizebox{\linewidth}{!}{ 
\begin{tabular}{l|ccc}
\textbf{Model} & \textbf{High-level} & \textbf{Mid-level} & \textbf{Low-level} \\
\hline
\rowcolor{blue!10} 
\multicolumn{4}{c}{\textbf{\textit{Open-source Models}}} \\
LLaVA-OneVision-72B                & 0.250  & 0.192 & 0.196 \\
Qwen-2.5-VL-72B             & 0.500  & 0.167 & 0.565 \\
InternVL-3-78B              & 0.125 & 0.485 & 0.500 \\
\rowcolor{blue!10}
\multicolumn{4}{c}{\textbf{\textit{Close-source Models}}} \\
Gemini-2.5-Pro                      & 0.500  & 0.218 & 0.208 \\
ChatGPT-4                       & 0.375 & 0.154 & 0.000 \\
Gemini-2.5-Flash            & 0.50  & 0.359 & 0.578 \\

\rowcolor{blue!10}
\multicolumn{4}{c}{\textbf{\textit{Qwen2.5-VL-3B}}} \\
\cellcolor{red!10}Qwen2.5-VL-3B               & 0.375 & 0.244 & 0.375 \\
\checkmark                   & 0.675 & 0.680 & 0.746 \\
\checkmark +SI                      & 0.875 & 0.795 & 0.750 \\
\checkmark +2D VP                & 1.000 & 0.603 & 0.958 \\
\checkmark +3D bbox                    & 1.000 & 0.769 & 1.000 \\
\checkmark +2D VP +3D bbox      & 1.000 & 0.820 & 1.000 \\
\checkmark +2D VP +3D bbox (All)      & 0.875 & 0.782 & 0.917 \\

\rowcolor{blue!10}
\multicolumn{4}{c}{\textbf{\textit{InternVL-3.5-2B}}} \\
\hline
\cellcolor{red!10}InternVL-3.5-2B           & 0.750  & 0.449 & 0.458 \\
\checkmark         & 0.875 & 0.731 & 0.746 \\
\checkmark +SI                       & 1.000  & 0.756 & 0.786 \\
\checkmark +2D VP               & 1.000  & 0.769 & 0.832 \\
\checkmark +3D bbox         & 0.875 & 0.744 & 0.917 \\
\checkmark +2D VP +3D bbox      & 1.000  & 0.846 & 1.000 \\
\checkmark +2D VP +3D bbox (All)      & 0.500  & 0.743 & 0.833 \\
\hline
\end{tabular}
}

\label{tab:longtail}
\end{wraptable}

\textbf{Cross-Domain Analysis.}
Three consistent trends emerge across all scene categories. 
\textit{First}, SpatialVLM is the strongest open-source model overall, 
frequently matching or surpassing closed-source models on 
geometry-sensitive tasks, attributable to its spatially grounded 
training pipeline. 
\textit{Second}, Gemini models demonstrate superior capability on 
high-level compositional reasoning across all three domains, suggesting 
stronger language-grounded spatial inference. 
\textit{Third}, all models degrade substantially on object-centric 
numerical estimation, revealing that precise metric reasoning from 
monocular single-object views remains an open challenge, one that 
MonoSR is uniquely positioned to benchmark.

\begin{figure*}[ht!]
    \captionsetup{type=table}
    \caption{
Impact of auxiliary information on monocular spatial reasoning across Indoor, Outdoor, and Object-Centric settings. \uline{Some unreasonable question types are filtered out depending on the scenario.} Dark blue and orange indicate the best result, while light colours indicate the second-best result.
}
    
    \centering
    \resizebox{\textwidth}{!}{
    \large 
    \begin{tabular}{ll|c|ccccc|ccccc|cccccccccccc}
    & & \textbf{Overall} &
    \rotatebox{75}{SR(MC4)} &
    \rotatebox{75}{SR(Y/N)} &
    \rotatebox{75}{Dist(Num.)} &
    \rotatebox{75}{Dim(Y/N)} &
    \rotatebox{75}{Size (Num.)} &
    \rotatebox{75}{OJ(MC3)} &
    \rotatebox{75}{OJ(Y/N)} &
    \rotatebox{75}{PR(MC4)} &
    \rotatebox{75}{PR(Y/N)} &
    \rotatebox{75}{OG(MC4)} &
    \rotatebox{75}{High SR(MC4)} &
    \rotatebox{75}{High SR(Y/N)} &
    \rotatebox{75}{High Dist(Num.)} &
    \rotatebox{75}{High Dim(Y/N)} &
    \rotatebox{75}{High Size (Num.)} &
    \rotatebox{75}{High OJ(MC3)} &
    \rotatebox{75}{High OJ(Y/N)} &
    \rotatebox{75}{High PR(MC4)} &
    \rotatebox{75}{High PR(Y/N)} &
    \rotatebox{75}{High OG(MC4)}
    \\     
    \textbf{FT} & \textbf{Aux Info} & &
    \multicolumn{5}{c}{\cellcolor{red!10}Low-level} &
    \multicolumn{5}{c}{\cellcolor{orange!10}Mid-level} &
    \multicolumn{10}{c}{\cellcolor{yellow!10}High-level} \\
    \hline

    \rowcolor{navyblue!10}
    \multicolumn{23}{c}{\textbf{\textit{Indoor Scene}}} \\
    
    \multicolumn{2}{l|}{\cellcolor{red!10 } QWen2.5-VL-3B} 
         & 0.283 
         & 0.312 & 0.495 & 0.063 & 0.440 & 0.074
         & 0.286 & 0.164 & 0.247 & 0.491 & 0.241 
         & 0.330 & 0.330 & 0.025 & 0.423 & 0.041 &
           0.269 & \cellcolor{navyblue!20}0.520 & 0.256 & 0.441 & 0.207 \\
    \checkmark & 
         & 0.494 
         & 0.700 & 0.684 & 0.230 & 0.724 & 0.160
         & 0.910 & 0.932 & 0.407 & 0.568 & 0.334 
         & 0.450 & 0.490 & 0.170 & 0.660 & 0.154 &
           0.904 & 0.400 & 0.250 & 0.540 & 0.220 \\
    \checkmark & + SI
         & 0.566 
         & 0.691 & 0.700 & 0.264 & 0.754 & 0.200
         & 0.947 & 0.961 & 0.404 & 0.535 & 0.757
         & 0.560 & 0.550 & 0.214 & 0.680 & 0.207 &
           0.950 &\cellcolor{navyblue!50} 0.550 &0.190 & 0.540 & 0.660 \\
    \checkmark & + 2D VP
         & 0.585 
         & 0.790 & 0.860 & 0.296 & 0.824 & 0.258
         & 0.846 & 0.867 & 0.427 & 0.620 & 0.758
         & 0.739 & 0.650 & 0.193 & 0.621 & 0.177 &
           0.955 & 0.422 &0.220 & 0.540 & 0.642 \\
    \checkmark & + 3D bbox
         & 0.722 
         &\cellcolor{navyblue!20} 0.914 & 0.776 &\cellcolor{navyblue!50} 1.000 & 0.808 &\cellcolor{navyblue!50} 1.000
         & 0.939 & 0.807 & \cellcolor{navyblue!20}0.469 & 0.598 & \cellcolor{navyblue!20}0.915
         & \cellcolor{navyblue!20}0.780 & 0.580 & 0.644 & 0.510 & 0.863 &
           0.937 & 0.392 &0.252 & 0.586 & 0.671 \\
    \checkmark & + 2D VP + 3D bbox
         &\cellcolor{orange!50}  0.798 
         &\cellcolor{navyblue!50} 0.940 & \cellcolor{navyblue!50}0.914 & \cellcolor{navyblue!50}1.000 & 0.890 &\cellcolor{navyblue!50} 1.000
         & 0.923 &\cellcolor{navyblue!20} 0.962 & 0.447 & 0.620 & \cellcolor{navyblue!50}0.944
         & \cellcolor{navyblue!50}0.890 &\cellcolor{navyblue!50} 0.895 & \cellcolor{navyblue!50}0.800 & \cellcolor{navyblue!20}0.780 & \cellcolor{navyblue!20}0.962 &
           0.950 & 0.350 &\cellcolor{navyblue!20}0.350 &\cellcolor{navyblue!50} 0.647 & \cellcolor{navyblue!20}0.700 \\
    \checkmark & + 2D VP + 3D bbox (All) 
         &\cellcolor{orange!15}  0.767 
         & 0.810 & 0.860 & \cellcolor{navyblue!50}1.000 & \cellcolor{navyblue!50}0.900 & \cellcolor{navyblue!50}1.000
         &\cellcolor{navyblue!50} 0.956 &\cellcolor{navyblue!50} 0.968 & \cellcolor{navyblue!50}0.497 &\cellcolor{navyblue!50} 0.644 & 0.757
         & 0.700 & 0.650 &\cellcolor{navyblue!20} 0.730 & \cellcolor{navyblue!50}0.840 & \cellcolor{navyblue!50}0.982 &
          \cellcolor{navyblue!20}0.956 & 0.440 &\cellcolor{navyblue!50}0.364 & 0.580 & \cellcolor{navyblue!50}0.710 \\
    \hline
    \multicolumn{2}{l|}{\cellcolor{red!10 } InternVL-3.5-2B} 
         & 0.301 
         & 0.275 & 0.272 & 0.047 & 0.152 & 0.044
         & 0.332 & 0.516 & 0.301 & 0.502 & 0.368 
         & 0.303 & 0.460 & 0.027 & 0.310 & 0.000 &
           0.540 & 0.560 & 0.251 & 0.470 & 0.284 \\
    \checkmark & 
         & 0.483 
         & 0.659 & 0.556 & 0.192 & 0.687 & 0.142
         & 0.620 & 0.604 & 0.387 & 0.554 & 0.654 
         & 0.550 & 0.540 & 0.112 & 0.546 & 0.162 &
           0.798 & 0.610 & 0.246 & 0.460 & 0.590 \\
    \checkmark & + SI
         & 0.515 
         & 0.691 & 0.752 & 0.210 & 0.730 & 0.140
         & 0.640 & 0.628 & 0.347 & 0.622 & 0.589
         & 0.560 & 0.600 & 0.110 & 0.690 & 0.180 &
           0.831 & 0.650 & 0.206 & 0.560 & 0.560 \\
    \checkmark & + 2D VP
         & 0.507
         & 0.617 & 0.620 & 0.180 & 0.660 & 0.267
         & 0.887 & 0.765 & 0.398 & 0.611 & 0.589
         & 0.500 & 0.620 & 0.070 & 0.660 & 0.023 &
           0.770 & 0.620 & 0.231 & 0.500 & 0.550 \\
    \checkmark & + 3D bbox
         &\cellcolor{orange!15} 0.724
         &\cellcolor{navyblue!20} 0.831 & 0.816 &\cellcolor{navyblue!20} 0.762 & 0.758 & 0.865
         &\cellcolor{navyblue!50} 0.946 & 0.933 &\cellcolor{navyblue!20} 0.486 & 0.559 & \cellcolor{navyblue!20}0.762
         & \cellcolor{navyblue!20}0.690 &\cellcolor{navyblue!50} 0.730 & 0.630 &\cellcolor{navyblue!20} 0.780 & 0.875 &
         \cellcolor{navyblue!20}  0.932 & 0.550 & 0.223 & \cellcolor{navyblue!20}0.570 &\cellcolor{navyblue!20} 0.780 \\
    \checkmark & + 2D VP + 3D bbox
         &\cellcolor{orange!50} 0.794 
         & \cellcolor{navyblue!50}0.882 &\cellcolor{navyblue!50} 0.932 &\cellcolor{navyblue!50} 1.000 &\cellcolor{navyblue!50} 0.936 & \cellcolor{navyblue!50}0.992
         &\cellcolor{navyblue!20}  0.944 &\cellcolor{navyblue!50} 0.960 & \cellcolor{navyblue!50}0.487 &\cellcolor{navyblue!50} 0.682 &\cellcolor{navyblue!50} 0.813
         & \cellcolor{navyblue!50}0.720 &\cellcolor{navyblue!20} 0.710 & \cellcolor{navyblue!50}0.701 &\cellcolor{navyblue!50} 0.810 &\cellcolor{navyblue!50} 0.978 &
           \cellcolor{navyblue!50}0.950 &\cellcolor{navyblue!50} 0.660 &\cellcolor{navyblue!20}0.332 & \cellcolor{navyblue!50}0.580 &\cellcolor{navyblue!50} 0.820 \\
    \checkmark & + 2D VP + 3D bbox (All) 
         & 0.721 
         & 0.752 &\cellcolor{navyblue!20} 0.829 &\cellcolor{navyblue!20} 0.762 &\cellcolor{navyblue!20} 0.853 &\cellcolor{navyblue!20} 0.935
         &0.906 &\cellcolor{navyblue!20}  0.956 & 0.464 &\cellcolor{navyblue!20} 0.626 & 0.635
         & 0.650 & 0.680 &\cellcolor{navyblue!20} 0.650 &\cellcolor{navyblue!20} 0.780 & \cellcolor{navyblue!20}0.904 &
          0.920 &\cellcolor{navyblue!20} 0.590 &\cellcolor{navyblue!50}0.351 & 0.530 & 0.650 \\
    \hline

    \rowcolor{navyblue!10}
    \multicolumn{23}{c}{\textbf{\textit{Outdoor Scene}}} \\
    
    \multicolumn{2}{l|}{\cellcolor{red!10 } QWen2.5-VL-3B} 
        & 0.302 
        & 0.385 & 0.474 &0.039 & $\times$ &$\times$ 
        & $\times$ & $\times$ & 0.241 & 0.504 & 0.250 
        & 0.304 & 0.430 & 0.023 & $\times$ & $\times$ &
          $\times$ & $\times$ & 0.280 & 0.430 & 0.261 \\
    \checkmark & 
         & 0.442 
         & 0.621 & 0.678 & 0.110 & $\times$ & $\times$ 
         & $\times$ & $\times$ & 0.410 & 0.562 & 0.354 
         &0.670 & 0.641 & 0.110 & $\times$ & $\times$ &
           $\times$ & $\times$ & 0.380 & 0.520 & 0.250 \\
    \checkmark & + SI
         & 0.558 
         & 0.700 & 0.688 & 0.220 & $\times$ & $\times$ 
         & $\times$ & $\times$ & 0.757 & 0.561 & 0.663 
         &0.630 & 0.620 & 0.190 & $\times$ & $\times$ &
           $\times$ & $\times$ &\cellcolor{navyblue!20} 0.470 &\cellcolor{navyblue!50} 0.640 & 0.560 \\
    \checkmark & + 2D VP
         & 0.576 
         & 0.762 & 0.760 & 0.180 & $\times$ & $\times$ 
         & $\times$ & $\times$ & 0.758 &\cellcolor{navyblue!20} 0.610 & 0.642 
         &\cellcolor{navyblue!20}0.733 & 0.658 & 0.222 & $\times$ & $\times$ &
           $\times$ & $\times$ & 0.419 &\cellcolor{navyblue!20} 0.579 &\cellcolor{navyblue!20} 0.584 \\
    \checkmark & + 3D bbox
         & 0.723 
         &\cellcolor{navyblue!20} 0.809 &\cellcolor{navyblue!20} 0.840 & \cellcolor{navyblue!20}0.998 & $\times$ & $\times$ 
         & $\times$ & $\times$ &\cellcolor{navyblue!20}0.915 & 0.574 & \cellcolor{navyblue!20}0.786 
         &0.727 & 0.667 & 0.732 & $\times$ & $\times$ &
           $\times$ & $\times$ & 0.426 & 0.509 &\cellcolor{navyblue!50} 0.690 \\
    \checkmark & + 2D VP + 3D bbox
         &\cellcolor{orange!50} 0.768 
         & \cellcolor{navyblue!50}0.871 &\cellcolor{navyblue!50} 0.907 & \cellcolor{navyblue!50}1.000 & $\times$ & $\times$ 
         & $\times$ & $\times$ & \cellcolor{navyblue!50}0.944 & \cellcolor{navyblue!50}0.618 & \cellcolor{navyblue!50}0.903
         &0.670 &\cellcolor{navyblue!50}0.895 &\cellcolor{navyblue!20} 0.828 & $\times$ & $\times$ &
           $\times$ & $\times$ & 0.450 &\cellcolor{navyblue!50} 0.640 & 0.487 \\
    \checkmark & + 2D VP + 3D bbox (All)          
         & \cellcolor{orange!15}0.732 
         & 0.760 & 0.810 & 0.980 & $\times$ & $\times$ 
         & $\times$ & $\times$ &  0.800& 0.604 & 0.726 
         &\cellcolor{navyblue!50}0.734 & \cellcolor{navyblue!20}0.770 & \cellcolor{navyblue!50}0.980 & $\times$ & $\times$ &
           $\times$ & $\times$ &\cellcolor{navyblue!50} 0.490 & 0.560 & 0.567 \\
    \hline
    \multicolumn{2}{l|}{\cellcolor{red!10 } InternVL-3.5-2B} 
        & 0.294 
        & 0.248 & 0.340 &0.017 & $\times$ &$\times$ 
        & $\times$ & $\times$ & 0.283 & 0.472 & 0.330 
        & 0.370 & 0.470 & 0.015 & $\times$ & $\times$ &
          $\times$ & $\times$ & 0.270 & 0.410 & 0.300 \\
    \checkmark & 
         & 0.462 
         & 0.716 & 0.558 & 0.132 & $\times$ & $\times$ 
         & $\times$ & $\times$ & 0.414 & 0.541 & 0.602 
         &0.650 & 0.550 & 0.052 & $\times$ & $\times$ &
           $\times$ & $\times$ & 0.370 & 0.470 & 0.490 \\
    \checkmark & + SI
         & 0.491 
         & 0.724 & 0.689 & 0.120 & $\times$ & $\times$ 
         & $\times$ & $\times$ & 0.420 & 0.522 & 0.589 
         &0.630 & 0.590 & 0.120 & $\times$ & $\times$ &
           $\times$ & $\times$ & 0.380 & 0.570 & 0.540 \\
    \checkmark & + 2D VP
         & 0.461 
         & 0.701 & 0.633 & 0.103 & $\times$ & $\times$ 
         & $\times$ & $\times$ & 0.392 & 0.540 & 0.589 
         &0.600 & 0.560 & 0.090 & $\times$ & $\times$ &
           $\times$ & $\times$ & 0.370 &0.440 &0.510 \\
    \checkmark & + 3D bbox
         &\cellcolor{orange!15} 0.740 
         &\cellcolor{navyblue!50} 0.885 & 0.762 & 0.930 & $\times$ & $\times$ 
         & $\times$ & $\times$ &0.753 & 0.599 &\cellcolor{navyblue!50} 0.729 
         &\cellcolor{navyblue!20}0.805 &\cellcolor{navyblue!20} 0.728 & 0.820 & $\times$ & $\times$ &
           $\times$ & $\times$ &\cellcolor{navyblue!50} 0.516 & 0.584 & \cellcolor{navyblue!20}0.768 \\
    \checkmark & + 2D VP + 3D bbox
         & \cellcolor{orange!50}0.773 
         &\cellcolor{navyblue!20} 0.882 &\cellcolor{navyblue!50} 0.858 & \cellcolor{navyblue!50}1.000 & $\times$ & $\times$ 
         & $\times$ & $\times$ &\cellcolor{navyblue!50} 0.862 &\cellcolor{navyblue!50} 0.681 &\cellcolor{navyblue!20} 0.704
         &\cellcolor{navyblue!50}0.815 &\cellcolor{navyblue!50}0.750 & \cellcolor{navyblue!20}0.841 & $\times$ & $\times$ &
           $\times$ & $\times$ &\cellcolor{navyblue!20} 0.462 &\cellcolor{navyblue!20} 0.594 &\cellcolor{navyblue!50} 0.822 \\
    \checkmark & + 2D VP + 3D bbox (All)          
         & 0.721 
         & 0.781 & \cellcolor{navyblue!20}0.787 &\cellcolor{navyblue!20} 0.930 & $\times$ & $\times$ 
         & $\times$ & $\times$ &\cellcolor{navyblue!20}  0.820&\cellcolor{navyblue!20} 0.643 & 0.635 
         &0.741 &\cellcolor{navyblue!50} 0.750 &\cellcolor{navyblue!50} 0.860 & $\times$ & $\times$ &
           $\times$ & $\times$ & 0.426 &\cellcolor{navyblue!50} 0.604 & 0.680 \\
    \hline

    \rowcolor{navyblue!10}
    \multicolumn{23}{c}{\textbf{\textit{Object-Centric Scene}}} \\
    
    \multicolumn{2}{l|}{\cellcolor{red!10 } QWen2.5-VL-3B}
        & 0.052 
        & $\times$ & $\times$ & 0.064& $\times$ & 0.057
        & $\times$ & $\times$
        & $\times$ & $\times$ & $\times$
        & $\times$ & $\times$ & 0.043 & $\times$ & 0.045 
        & $\times$ & $\times$ & $\times$ & $\times$ & $\times$ \\
    \checkmark &
        & 0.335 
        & $\times$ & $\times$ & 0.320& $\times$ & 0.400
        & $\times$ & $\times$
        & $\times$ & $\times$ & $\times$
        & $\times$ & $\times$ & 0.340 & $\times$ & 0.280 
        & $\times$ & $\times$ & $\times$ & $\times$ & $\times$ \\
    \checkmark & + SI
        & 0.381 
        & $\times$ & $\times$ & 0.390& $\times$ & 0.446 
        & $\times$ & $\times$
        & $\times$ & $\times$ & $\times$
        & $\times$ & $\times$ &0.363 & $\times$ &  0.323
        & $\times$ & $\times$ & $\times$ & $\times$ & $\times$ \\
    \checkmark & + 2D VP
        & 0.437 
        & $\times$ & $\times$ &\cellcolor{navyblue!20} 0.457 & $\times$ & 0.467 
        & $\times$ & $\times$
        & $\times$ & $\times$ & $\times$
        & $\times$ & $\times$ & 0.375 & $\times$ & 0.450
        & $\times$ & $\times$ & $\times$ & $\times$ & $\times$ \\ 
    \checkmark & + 3D bbox
        & 0.986
        & $\times$ & $\times$ & \cellcolor{navyblue!50}1.000 & $\times$ &\cellcolor{navyblue!20} 0.994
        & $\times$ & $\times$
        & $\times$ & $\times$ & $\times$
        & $\times$ & $\times$ &\cellcolor{navyblue!50} 1.000 & $\times$ & 0.948 
        & $\times$ & $\times$ & $\times$ & $\times$ & $\times$ \\
    \checkmark & + 2D VP + 3D bbox
        &\cellcolor{orange!50} 0.996 
        & $\times$ & $\times$ &\cellcolor{navyblue!50} 1.000& $\times$ & 0.983 
        & $\times$ & $\times$
        & $\times$ & $\times$ & $\times$
        & $\times$ & $\times$ &\cellcolor{navyblue!50} 1.000 & $\times$ &\cellcolor{navyblue!50} 1.000 
        & $\times$ & $\times$ & $\times$ & $\times$ & $\times$ \\
    \checkmark & + 2D VP + 3D bbox (All)
        & \cellcolor{orange!15}0.988 
        & $\times$ & $\times$ &\cellcolor{navyblue!50} 1.000& $\times$ &\cellcolor{navyblue!50} 1.000 
        & $\times$ & $\times$
        & $\times$ & $\times$ & $\times$
        & $\times$ & $\times$ &\cellcolor{navyblue!20} 0.980 & $\times$ &\cellcolor{navyblue!20} 0.970 
        & $\times$ & $\times$ & $\times$ & $\times$ & $\times$ \\
    \hline
     \multicolumn{2}{l|}{\cellcolor{red!10 } InternVL-3.5-2B}
        & 0.033 
        & $\times$ & $\times$ & 0.041& $\times$ & 0.024
        & $\times$ & $\times$
        & $\times$ & $\times$ & $\times$
        & $\times$ & $\times$ & 0.025 & $\times$ & 0.041 
        & $\times$ & $\times$ & $\times$ & $\times$ & $\times$ \\
    \checkmark &
        & 0.299 
        & $\times$ & $\times$ & 0.257& $\times$ & 0.424
        & $\times$ & $\times$
        & $\times$ & $\times$ & $\times$
        & $\times$ & $\times$ & 0.210 & $\times$ & 0.304 
        & $\times$ & $\times$ & $\times$ & $\times$ & $\times$ \\
    \checkmark & + SI
        & 0.349 
        & $\times$ & $\times$ & 0.250& $\times$ & 0.460 
        & $\times$ & $\times$
        & $\times$ & $\times$ & $\times$
        & $\times$ & $\times$ &0.363 & $\times$ &  0.323
        & $\times$ & $\times$ & $\times$ & $\times$ & $\times$ \\
    \checkmark & + 2D VP
        & 0.280 
        & $\times$ & $\times$ & 0.270 & $\times$ & 0.380 
        & $\times$ & $\times$
        & $\times$ & $\times$ & $\times$
        & $\times$ & $\times$ & 0.120 & $\times$ & 0.350
        & $\times$ & $\times$ & $\times$ & $\times$ & $\times$ \\ 
    \checkmark & + 3D bbox
        & 0.930
        & $\times$ & $\times$ & 0.920 & $\times$ &\cellcolor{navyblue!20}0.970
        & $\times$ & $\times$
        & $\times$ & $\times$ & $\times$
        & $\times$ & $\times$ & 0.850 & $\times$ & \cellcolor{navyblue!20}0.980 
        & $\times$ & $\times$ & $\times$ & $\times$ & $\times$ \\
    \checkmark & + 2D VP + 3D bbox
        &\cellcolor{orange!50} 0.976 
        & $\times$ & $\times$ &\cellcolor{navyblue!50} 1.000& $\times$ &\cellcolor{navyblue!50} 0.990 
        & $\times$ & $\times$
        & $\times$ & $\times$ & $\times$
        & $\times$ & $\times$ & \cellcolor{navyblue!50}0.925 &   $\times$ & \cellcolor{navyblue!50}0.988 
        & $\times$ & $\times$ & $\times$ & $\times$ & $\times$ \\
    \checkmark & + 2D VP + 3D bbox (All)
        &\cellcolor{orange!15} 0.934 
        & $\times$ & $\times$ &\cellcolor{navyblue!20} 0.950& $\times$ & 0.930 
        & $\times$ & $\times$
        & $\times$ & $\times$ & $\times$
        & $\times$ & $\times$ &\cellcolor{navyblue!20} 0.904 & $\times$ & 0.950 
        & $\times$ & $\times$ & $\times$ & $\times$ & $\times$ \\
    \hline
    \end{tabular}
 }

    \label{table:input_inf}
\end{figure*}

\subsection{Impact of Auxiliary Information}
Tab.~\ref{table:input_inf} presents the performance improvements attained by incorporating different types of auxiliary information across the Indoor, Outdoor, and Object-Centric settings. Details of these auxiliary inputs are provided in Sec.~\ref{sec:aux}. Throughout this study, we fine-tune two architecturally distinct 
backbones, Qwen2.5-VL-3B\cite{Bai2023QwenVL} and InternVL-3.5-2B\cite{wang2025internvl35}, under identical 
auxiliary input configurations to verify that observed trends 
generalize across model families rather than being architecture-specific.

We first observe that fine-tuning on MonoSR alone, without any 
auxiliary input, already yields substantial gains over the frozen 
baseline: Qwen2.5-VL-3B improves from 0.283 to 0.494 on Indoor 
and from 0.302 to 0.442 on Outdoor; InternVL-3.5-2B shows similar 
trends (0.301→0.483 Indoor, 0.294→0.462 Outdoor). This consistent 
improvement confirms that MonoSR provides effective spatial 
supervision that activates latent reasoning capabilities already 
present in the base models.

Building on this, introducing global scene information (SI), which explicitly indicates the scene type of the input image, leads to additional improvements across both low- and mid-level tasks. This suggests that high-level contextual priors help the model better calibrate its spatial predictions.

Adding 2D visual prompts (2D VP) further amplifies these gains, especially for mid-level indoor tasks that rely heavily on semantic grounding and local spatial relations. This highlights that, even for 3D spatial reasoning, localized 2D spatial cues remain highly effective in anchoring the model’s understanding of object layouts and relational structure.
The largest improvements come from incorporating 3D bounding boxes (3D bbox). With access to explicit 3D geometric structure, the model achieves near-perfect accuracy on tasks involving object arrangement, spatial proximity, and precise geometric comparison, highlighting the inherent limitations of purely image-based reasoning.

The combination of 2D visual prompts + 3D bounding boxes yields the most robust and stable improvements across all difficulty levels. Their complementary contributions, with semantic grounding provided by 2D prompts and geometric structure captured by 3D bounding boxes, provide reliable spatial cues that substantially enhance performance.
 Although Outdoor scenes remain more challenging due to diverse environments and clutter, the same pattern persists: multimodal fusion of 2D and 3D signals consistently delivers strong gains despite the inherent complexity.
Results from the Object-Centric setting further reinforce these observations. Even when each image contains only a single object, the model still benefits substantially from explicit 3D structural guidance. This confirms that the primary bottleneck is not object recognition, but the recovery of fine-grained spatial attributes from monocular inputs.

Finally, we provide the model with 2D visual prompts together with 
3D bounding boxes for all detected objects in the scene. This holistic 
configuration yields slightly worse performance compared to the 
targeted variant, for instance, Indoor Overall drops from 0.798 to 
0.767 (Qwen2.5-VL-3B) and from 0.794 to 0.721 (InternVL-3.5-2B), 
with similar degradation observed in Outdoor and Object-Centric 
settings. This indicates that performance gains arise only from 
relevant auxiliary information: introducing excessive or irrelevant 
cues can distract the model and degrade its spatial reasoning ability, 
suggesting that future monocular perception modules should focus on 
targeted rather than exhaustive spatial signal injection.

\subsection{Long-tail Evaluation}
We emphasize the open-world nature of MonoSR, which encompasses a 
substantially broader spectrum of object categories than prior 
benchmarks. To evaluate generalization to rare categories, we report 
results on long-tail classes (bottom 20\% by frequency) in 
Tab.~\ref{tab:longtail}. All values are accuracy scores in $[0,1]$. 
Models fine-tuned on MonoSR consistently outperform their untuned 
baselines across all three reasoning levels, confirming that MonoSR's 
open-world coverage enhances generalization to infrequent objects 
rather than overfitting to frequent ones.

\subsection{Fine-grained Analysis}

To better understand model performance on \textit{Situational Reasoning}, Table~\ref{tab:situational_breakdown} reports results across four scenario types: \textit{Compliance} (C.), \textit{Planning} (P.), \textit{Safety} (S.), and \textit{Navigation} (N.). Planning consistently remains the most challenging scenario.

\begin{table}[t]
\centering
\caption{Performance (\%) across four situational reasoning scenarios. C., P., S., and N. denote \textit{Compliance}, \textit{Planning}, \textit{Safety}, and \textit{Navigation}, respectively.}
\label{tab:situational_breakdown}
\setlength{\tabcolsep}{8pt}
\renewcommand{\arraystretch}{1.15}
\resizebox{0.8\columnwidth}{!}{
\begin{tabular}{lccccc}
\toprule
\textbf{Model} & \textbf{C.} & \textbf{P.} & \textbf{S.} & \textbf{N.} & \textbf{Avg.} \\
\midrule
InternVL3-72B & 43.32 & 33.33 & 40.55 & 45.45 & 41.24 \\
Gemini-2.5 Flash & 45.54 & 47.92 & 43.55 & 50.00 & 44.46 \\
\bottomrule
\end{tabular}
}
\vspace{-3mm}
\end{table}

\subsection{Dataset Visualization}
Finally, we present visualization results of MonoSR in Fig.~\ref{fig:visualization}, illustrating the richness and diversity of the dataset across indoor, outdoor, and object-centric scenarios. Each example contains a paired image, question, and ground-truth answer, covering a wide spectrum of spatial reasoning tasks. As shown in the figure, MonoSR includes complex multi-object interactions, cross-view geometric reasoning, and fine-grained metric queries that require precise spatial understanding beyond high-level semantics.
These examples highlight the challenging nature of MonoSR.

\section{Limitations \& Future Work}

MonoSR currently covers static scenes and does not address 
reasoning under viewpoint uncertainty, \eg, predicting spatial 
relationships from novel or occluded perspectives beyond the 
geometric transformations used in Perspective-Aware Imagination 
tasks. Closing this gap may require integrating neural rendering 
or scene completion as a data generation backbone. Additionally, 
while the auxiliary information study (Sec.~\ref{sec:aux}) 
establishes 3D bounding boxes as a geometric upper bound, 
bridging this oracle gap in a practical monocular system remains 
an open challenge. We plan to integrate metric depth estimation 
and monocular 3D detection modules directly into VLM architectures, 
using our auxiliary analysis as a blueprint for which signals 
matter most. Finally, RL-based post-training~\cite{Shao2024DeepSeekMath} 
offers a promising path toward improving spatial reasoning without 
additional labeled data, which we plan to explore using MonoSR's 
training split.

\section{Conclusion}
\label{sec:conclusion}

We present MonoSR, the large-scale dataset purpose-built for 
open-world monocular spatial reasoning, with guaranteed single-view 
answerability across indoor, outdoor, and object-centric domains. 
Through comprehensive evaluation of seven state-of-the-art VLMs, 
we show that monocular spatial reasoning, particularly metric 
estimation from object-centric views, remains far from solved, 
even for the strongest closed-source systems. Our auxiliary 
information study further reveals that targeted geometric signals 
yield large gains while irrelevant cues actively degrade 
performance, pointing to selective monocular perception as the 
key design principle for future models. We hope MonoSR serves 
as a rigorous testbed and training resource that accelerates 
progress toward VLMs capable of the effortless spatial 
understanding humans perform from a single glance.

\section*{Acknowledgments} 
This work is supported by the Agency for Science, Technology and Research (A*STAR) under its Career Development Fund (Project No. H26-KSR0066) and is also supported by the Agency for Science, Technology and Research (A*STAR) under its MTC Programmatic Funds (Grant No. M23L7b0021).

The authors would like to sincerely thank the Program Chairs, Area Chairs, and Reviewers for the time and efforts devoted during the review process.

%
%
\bibliographystyle{splncs04}
\bibliography{main}
\end{document}